\pdfoutput=1

\documentclass[journal]{IEEEtran}
\ifCLASSINFOpdf
\else
\fi

\usepackage{algorithm}
\usepackage{algpseudocode}
\usepackage{amsmath}
\usepackage{graphics}
\usepackage{epsfig}
\usepackage{multirow}
\usepackage[square,sort,comma,numbers]{natbib}
\usepackage{booktabs}
\usepackage{amsfonts}
\usepackage{color}
\usepackage{array}

\usepackage{verbatim}
\newcommand{\tabincell}[2]{\begin{tabular}{@{}#1@{}}#2\end{tabular}}

\begin{document}
%
\title{ Quality-aware Unpaired Image-to-Image Translation}
%
%
%
\author{Lei Chen, Le Wu, Zhenzhen Hu, Meng Wang,~\IEEEmembership{Senior Member,~IEEE}

\thanks{ 
This work was supported in part by grants from the National Natural Science Foundation for Distinguished Young Scholars of China (Grant No. 61725203), the National Key Research and Development Program of China (Grant No. 2017YFB0803301), and the National Natural Science Foundation of China( Grant No. 61602147, 61732008, 61802104).

L.~Chen,  L.~Wu~(corresponding author), Z.~Hu, M.~ Wang are with the School of Computer and Information, Hefei University of Technology, Hefei, Anhui 230009, China.
Emails: \{chenlei.hfut, lewu.ustc, huzhen.ice@gmail.com, eric.mengwang\}@gmail.com.
}
}

\markboth{IEEE Transactions on Multimedia}%
{Shell \MakeLowercase{\textit{et al.}}: Bare Demo of IEEEtran.cls for IEEE Journals}
%



\maketitle

\begin{abstract}

Generative Adversarial Networks (GANs) have been widely used for the image-to-image translation task. While these models rely heavily on the labeled image pairs, recently some GAN variants have been proposed to tackle the unpaired image translation task. These models exploited supervision at the domain level with a reconstruction process for unpaired image translation. On the other hand, parallel works have shown that leveraging perceptual loss functions based on high level deep features could enhance the generated image quality. Nevertheless, as these GAN-based models either depended on the pretrained deep network structure or relied on the labeled image pairs, they could not be directly applied to the unpaired image translation task. Moreover, despite the improvement of the introduced perceptual losses from deep neural networks, few researchers have explored the possibility of improving the generated image quality from classical image quality measures. To tackle the above two challenges, in this paper, we propose a unified quality-aware GAN-based framework for unpaired image-to-image translation, where a quality-aware loss is explicitly incorporated by comparing each source image and the reconstructed image at the domain level. Specifically, we design two detailed implementations of the quality loss. The first method is based on a classical image quality assessment measure by defining a classical quality-aware loss to ensure similar quality score between an original image and the reconstructed image at the domain level. The second method proposes an adaptive deep network based loss that compares the high level content structure between each original image and its reconstructed image from the generator. Finally, extensive experimental results on many real-world datasets clearly show the quality improvement of our proposed framework, and the superiority of leveraging classical image quality measures for unpaired image translation compared to the deep network based model.

\end{abstract}


%
\IEEEpeerreviewmaketitle

\section{Introduction}

\IEEEPARstart{M}any real-world computer vision tasks, such as image segmentation, stylization and abstraction, could be treated as an image-to-image translation problem. This problem involves transforming an image from a source domain~(e.g., photo) to imitate the image in the target domain~(e.g., sketch)~\cite{yi2017dualgan,elgammal2017can,zhu2017unpaired,kim2017learning}. Since the seminal work of GANs by Goodfellow et al. at 2014~\cite{goodfellow2014generative}, GAN and their variants provide state-of-the-art solutions to the image-to-image translation task. Given an image pair with a source image and its corresponding image in the target domain, GANs learn an adversarial loss function that tries to maximize the discriminator to correctly classify if the generated image is real or fake in the target scene,  and simultaneously trains a generative model that tries to fake the discriminator.

Generally, these classical GAN-based models rely on labeled image pairs from the source domain and the target domain for image translation. Nevertheless, in the real-world, it is relatively easy to collect different images in the source domain and target domain separately, while acquiring such side-by-side matching pairs is time and labor consuming~\cite{he2016dual,wang2012movie2comics}. E.g., we could collect a set of images in the summer scene and a set of images in the winter scene. However, it is nontrivial to get a side-by-side pair of an image taken in the summer next to the same matching image that is taken in the winter. Thus, some unpaired GAN variants have been proposed to translate an image from one  domain to the remaining domain without any paired images, including DualGAN \cite{yi2017dualgan}, CycleGAN \cite{zhu2017unpaired}, and DiscoGAN \cite{kim2017learning}. Roughly, all these models shared a similar idea. While a normal GAN has only one generator and one discriminator that is trained on the labeled image pairs, these unpaired GANs exploit supervision at the domain set level. Specifically, for image translation from a domain U to a domain V, there are two GANs with one primal GAN with generator $G_U$ learns to translate an image from domain U to that in domain V, and a corresponding GAN  with generator $G_V$ that learns to invert the translated image in V to the original domain U with an additional reconstruction loss.

\begin{figure}
\begin{center}
\includegraphics[width=8cm,clip]{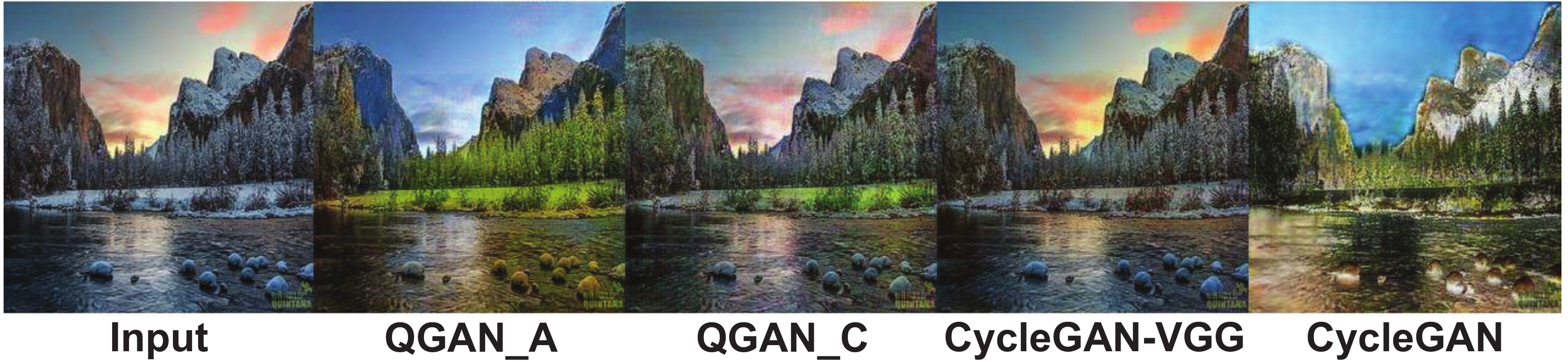}
\vspace{-0.2cm}
\caption{ An example of the winter-to-summer translation task. Given an input from the winter style, it is desired that the generated output images are visually qualified from the human perspective. Here, CycleGAN~\cite{zhu2017unpaired} is a state-of-the-art unpaired image-to-image translation model. CycleGAN-VGG combines CycleGAN and perceptual loss functions based on the pretrained VGG~\cite{ledig2017photo}. QGAN\_A and QGAN\_C are our proposed two quality-aware models.}
\label{Fig:sketch2photo1}
\end{center}
\vspace{-0.5cm}
\end{figure}

In these GAN variants for (unpaired) image translation task, the performance relies heavily on the designed optimization loss function.  All these GAN-based objective function includes an adversarial loss that alternates between identifying and faking. Besides, to ensure the similarity between an original image and its generated version, some (unpaired) GAN variants have introduced the pixel level loss in the modeling process~\cite{yi2017dualgan,zhu2017unpaired,kim2017learning,isola2017image},
e.g., the error between them with L1 or L2 distance loss~\cite{isola2017image}. Nevertheless, even though the pixel level matching consistency is high, people are not satisfied with the generated images as human visual system usually focus on the higher level abstractions of images for perceptual quality evaluation~\cite{ledig2017photo,chen2014quality}. Thus, some works focused on leveraging perceptual loss functions based on high level features of deep networks to enhance the generated image quality~\cite{wang2016generative,wang2018high}. E.g., Johnson et al. introduced perceptual loss functions that depended on high level image features extracted from a pretrained deep neural network~\cite{johnson2016perceptual}.
The newly added perceptual loss functions could generate quality-enhanced results for some specific image translation tasks. However, these pretrained deep networks are not optimized for the image-to-image translation task, not to mention the situation when the target image domain never appears in the pretrained network. For example, as shown in Fig.~\ref{Fig:sketch2photo1}, the fifth column is the results of an unpaired image translation model of CycleGAN~\cite{zhu2017unpaired}. We also show the results that combined CycleGAN~\cite{zhu2017unpaired} and the pretrained VGG for perceptual loss control in the fourth column. It is clearly observed that some patches in the fourth and fifth column are not summer and contain blurry structure. To tackle the suboptimal performance caused by the pretrained network, Wang et al. designed a perceptual adversarial loss that undergoes an adversarial training process for image translation~\cite{wang2018high}.  In the proposed model, the adversarial discriminator evaluates the perceptual loss between the generated image and the ground truth image with the discriminator network, achieving impressive results for image translation.


In this paper, we study the problem of generating high quality images for unpaired image-to-image translation. A natural idea is to combine the recent progress of unpaired image translation and the quality enhancement techniques for paired image translation. Some researchers proposed to introduce a content-based loss in the optimization function, where the content-based loss is constructed from pretrained deep networks between a source image and the corresponding generated image~\cite{johnson2016perceptual,ledig2017photo,zhang2018unreasonable,guo2018automatic}. 
These models could generate images with better quality than previous image translation models.
However, as the pretrained network parameters are not designed for image translation, simply transferring these parameters would lead to suboptimal quality-aware unpaired image translation task. Therefore, how to adaptively design the content-based quality loss to the unpaired image translation task to further enhance image quality still remains open. Moreover, image quality assessment is a classical topic that has been well studied in the community~\cite{wang2004image,xue2014gradient,zhang2011fsim}. As most image translation models relied on the deep features to model the perceptual losses for enhancing image quality, the question of whether the classical image quality measures could help to improve the quality of generated images is still under explored. To tackle the above two challenges, in this paper, we propose a unified quality-aware GAN-based framework for unpaired image-to-image translation. We extend the adversarial loss functions in unpaired image translation models and put emphasis on how to design quality-aware loss functions that could be applied to the image translation task without image pair information. Instead of comparing the quality between $u$ and the output of generator $G_U$ as $G_U(u)$, we model the quality  loss terms between $u$ and the reconstructed images of $u$~(i.e, $\hat{u}=G_V(G_U(u))$). Thus, the designed quality-aware loss functions are freed from the image pair constraints. Specifically, we introduce two detailed models to implement the quality loss. We first borrow the ideas of these classical quality assessment methods and propose a method that defines the quality loss between each original image and the reconstructed image with a classical image quality assessment measure. We then propose an adaptive content loss that combines the visual content structure loss from GANs for quality-aware unpaired image-to-image translation. The content loss adaptively captures the high level perceptual quality between the original images and the reconstructed images in the generators.

In summary, our paper makes the following contributions:

\begin{enumerate}
\renewcommand{\labelenumi}{(\theenumi)}
\setlength{\itemsep}{0pt}
\item  We point out that current image-to-image translation tasks relied on the labeled image pairs for designing
image quality based losses. To this end, we propose a unified quality-aware unpaired GAN-based image translation framework,  which relies on the quality losses between  $u$ and $\hat{u}=G_V(G_U(u))$.
\item  Under the proposed quality-aware framework, we design two detailed model implementations of the quality loss. The first proposed model introduces a classical quality assessment loss, and the second model combines a high level adaptive visual content structure loss in addition to the adversarial loss in GAN for modeling human perceptual quality evaluation.
\item 	We perform extensive experimental results on four real-world datasets. Extensive experimental results from both the quality assessment measures and the human opinion scores show that our proposed models improve the quality of the generated images. Also, we observe the superiority of the classical quality assessment loss
    compared to the high level content-based loss.
\end{enumerate}

\section{Related work}\label{sec:related}

The idea of image-to-image translation goes back at least to Gatys et al.'s neural algorithm of artistic style \cite{gatys2015neural}, which designed a neural algorithm to separate content and style and then recombined the two parts. Since the seminal work of GAN by Goodfellow et al. at 2014~\cite{goodfellow2014generative}, the recent image-to-image translation task has been tackled under the GAN framework. GANs learn an adversarial optimization function that maximizes the discriminator to correctly classify if the output image is real or fake, and simultaneously a generative model that minimizes the loss. Different GAN variants for the image-to-image task varied in the specific implementations of the discriminator and the generator. E.g., Isola et al. proposed a conditional adversarial network with generic loss function~\cite{isola2017image}  and Mao et al. presented a GAN variant that adopted the least square based loss function for the discriminator~\cite{mao2017least}. While these models relied on image pairs for the translation task, recently some unpaired image-to-image models have been proposed~\cite{yi2017dualgan,zhu2017unpaired,kim2017learning}. These models shared a similar idea by learning a primal GAN from a source domain to a target domain, and a dual process that transformed the generated images from the target domain to the source domain. Then, a pixel level reconstruction loss, such as mean squared error~(MSE)~\cite{dong2016image,dong2014learning}, is introduced between the original images and the reconstructed images. These models advanced previous works by loosing the inputs to unlabeled training data.

In fact, the performance of the image translation task relies heavily on the designed optimization function, which can effectively drive the network's learning, leading to a large impact on the performance of this task.
In the real world, the human visual system is quite subjective and human usually focus on the higher level abstractions for perceptual quality evaluation~\cite{zhang2011fsim,ledig2017photo,morrone1988feature}. Nevertheless, the pixel-wise loss functions suffered from the limitation of poorly reflecting the human visual experience, and thus typically induce  blurry parts. Luckily, \emph{C}onvolutional \emph{N}eural \emph{N}etworks~(CNN) have shown promising performance to automatically extract high level content structure information of images~\cite{lecun2015deep,schmidhuber2015deep,luo2016image}. Many works empirically validated that the higher layers of the CNN network capture the perceptual abstractions of images. Thus, many image processing related tasks, such as image resolution~\cite{ledig2017photo,liu2016robust}, style transfer~\cite{johnson2016perceptual,zhong2019camstyle} and unsupervised depth and motion estimation~\cite{yin2018geonet,godard2017unsupervised,li2018undeepvo,srinivasan2017learning}, are proposed to leverage the feature maps in CNNs as perceptual quality measures, and incorporated the perceptual quality into the optimization function to generate quality-aware images~\cite{johnson2016perceptual,ledig2017photo,zhang2018unreasonable,guo2018automatic}. Usually, the perceptual loss function~\cite{ledig2017photo} modeled the image features from a pretrained VGG network~\cite{simonyan2014very}. As these models do not need any labeled information for quality evaluation, they could be applied to the unpaired image translation task. Without confusion, in the following of this paper, we use ``CycleGAN-VGG'' to denote the unpaired image translation task that combined the loss of CycleGAN and the perceptual loss function from pretrained VGG network. However, since these pretrained networks are tailored to a specific dataset, it is suboptimal to directly transfer them to the image translation task. For example, as shown in Fig~\ref{Fig:sketch2photo1}, CycleGAN-VGG fails for the unpaired winter-to-summer translation as the pretrained network could not effectively capture the perceptual network configurations for winter and summer domain. To tackle the limitations of the pretrained network, recently researchers proposed a perceptual adversarial loss that undergoes an adversarial training process between the generator and the discriminator~\cite{wang2018perceptual}. The proposed model introduced an adaptive perceptual loss that automatically discovered the discrepancy between the generated image and the ground truth with higher layer based abstractions from deep networks.
To tackle the case in unpaired image translation, the authors use a small paragraph to illustrate how to extend this method to unpaired image translation. To avoid using the ground truth of labeled image pairs, instead of comparing the generated image and its ground truth, they proposed to calculate the discrepancy of mean features of these two domains. This proposed method showed better performance for paired image translation. However, as it is not dedicated for unpaired image translation, simply comparing the mean discrepancy between two domains would discard the characteristics of the currently generated image, leading to unsatisfactory performance. In summary, our work borrows the ideas of these previous works, and we advance in the following two aspects. Firstly, we would like to explore how to effectively improve the perceptual quality of the generated images in unpaired image translation task. Secondly, despite the breakthroughs of leveraging perceptual losses derived from deep neural networks, to the best of our knowledge, we are one of the first few attempts that explore the possibility of combining the classical quality assessment measures for unpaired image translation.


Our work is also closely related to the Image Quality Assessment~(IQA) measures. This direction aims to use computational models to measure the image quality that is consistent with human subjective evaluations. Generally speaking, current IQA techniques mainly follow two directions: the blind reference~(BR) and the full reference~(FR). BR evaluates image quality without any reference, and this kind of models usually designed some features for quality modeling~\cite{saad2010dct,gu2015using,mittal2013making,wu2017blind}.  However, as the overall quality of each domain varies~\cite{mittal2013making,wang2004image, moorthy2010two,lu2015rating}, these models either needed human labeled image quality values or were only applicable to a specific domain for quality evaluation. Different from BR, FR usually evaluates the visual quality of an image by comparing a generated image with the original image in the image-to-image translation task~\cite{wang2004image,zhang2011fsim}.  Instead by comparing the pixel level similarity such as peak signal-to-noise ratio (PSNR) and the mean-squared error (MSE) that directly operate on the pixel level of images, the FR methods show great success by designing the specific subjective features to simulate human visual evaluation. E.g., SSIM~\cite{wang2004image} proposed a complementary method for structural similarity. Based on the physiological and psychophysical evidences, FSIM~\cite{zhang2011fsim} emphasized the human visual system to understand the image based on the Fourier low frequency features of images. As the human visual system is adapted to structural information of images, GMSD~\cite{xue2014gradient} is proposed to use the gradient similarity based method  to measure image quality efficiently. With the development of deep neural networks, recently some methods of IQA have made preliminary attempts for automatically capturing image quality related features from  deep neural networks~\cite{kang2014convolutional,bosse2016deep,hou2015blind,sheikh2005no}. But all of this methods directly or indirectly required examples and corresponding human opinion scores,  which usually are expensive. In this work, we would like to borrow a classical FR method for designing a quality-aware unpaired image translation model.



\begin{figure*}[h]
\vspace{-0.2cm}
\includegraphics[width=1\textwidth]{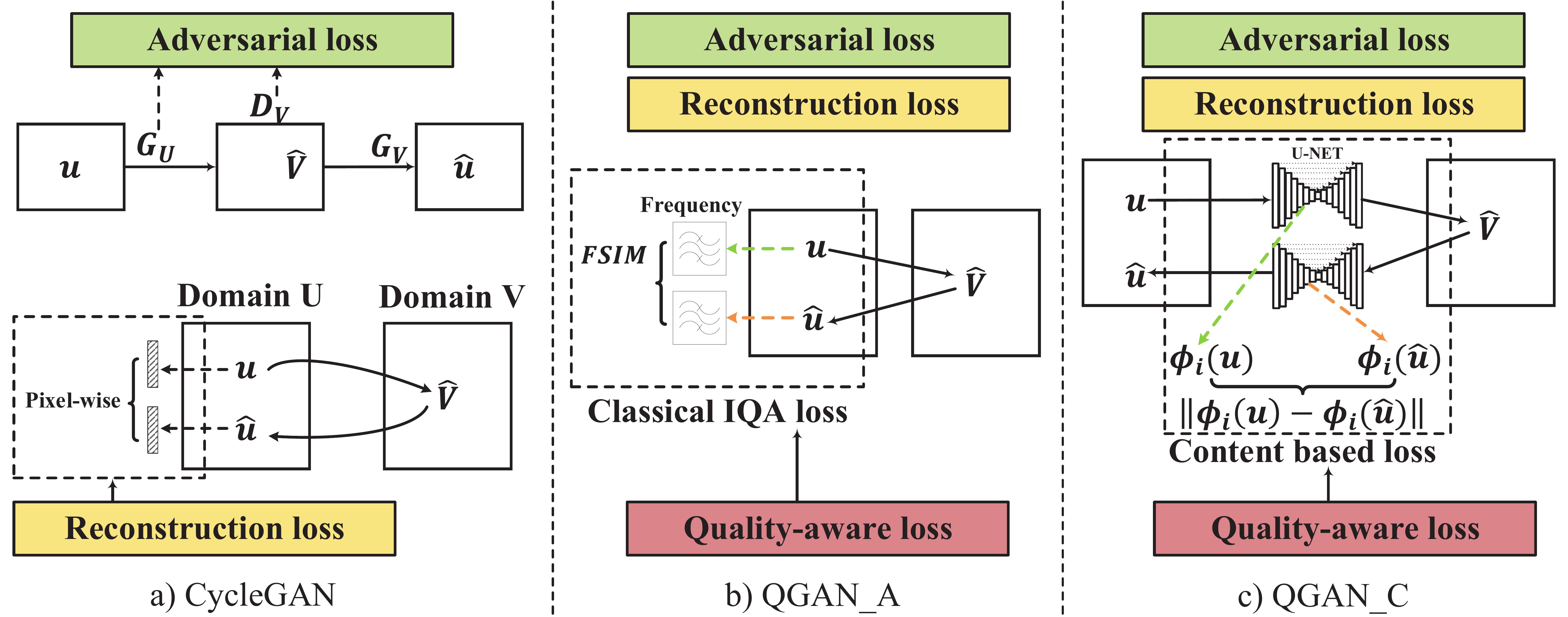}
\vspace{-0.7cm}
\caption{Our proposed Quality-aware GAN-based models for unpaired image-to-image translation. The leftmost part shows the
CycleGAN~\cite{zhu2017unpaired}, and the remaining two parts shows our proposed two detailed models under the QGAN framework. In the middle part, our proposed QGAN\_A model designs a quality loss term that is based on a classical image quality assessment model of FSIM. The QGAN\_C model designs an adaptive content loss that captures the high level perceptual structure of images from the GANs without any labeled image pairs.}
\label{Fig:quality_awareGANs}
\vspace{-0.5cm}
\end{figure*}

\section{Preliminaries}\label{sec:pre}
In this section, we introduce the key ideas of several recent state-of-the-art GAN-based  unpaired image-to-image translation models, including DualGAN \cite{yi2017dualgan}, CycleGAN \cite{zhu2017unpaired}, and DiscoGAN \cite{kim2017learning}. Since all these models share a similar idea, for ease of explanation, we take the CycleGAN \cite{zhu2017unpaired} as an example to show the key ideas of the  GAN-based  unpaired image-to-image translation models.

CycleGAN learns to translate an image $u$ from a source domain $U$~($u\in U$) to a target domain $V$  with a generator $G_U$ in the absence of paired examples, such that $G_U(u)$ is indistinguishable from the distribution of images in $V$. As this process is highly under constrained, CycleGAN couples this process with an inverse mapping $G_V: V\rightarrow U$. Correspondly, a cycle consistent loss, i.e., $u$ and $\hat{u}\!=\!G_V(G_U(u))$ is introduced for each image in the source domain.  Besides, there are two adversarial discriminators: $D_U$ and $D_V$. $D_U$ aims to distinguish between images in $U$ and the translated images with generator $G_V$. Similarly, $D_V$ is a discriminator that distinguishes between images in $V$ and the translated images with generator $G_U$. The closed loop which is made by the cycle consistent loss allows images from either domain to be translated and then reconstructed.  Given the above analysis, the objective
loss function in CycleGAN contains two terms: the adversarial loss~${L}_{GAN}(u,v)$ inherited from the GAN-based model that tries to match the distribution of the generated images to the real image distribution in that domain, and the reconstruction loss~${L}_R(u,v)$ to ensure the learned dual mappings are consistent with the images:

\vspace{-0.5cm}
\begin{equation}\label{eq:dualgan_loss}
\vspace{-0.25cm}
\begin{split}
L(u,v)=&L_{GAN}(u,v)+L_{R}(u,v),
\end{split}
\vspace{-0.1cm}
\end{equation}

\noindent where $u\subseteq U$, $v\subseteq V$.

In the above equation, for the adversarial loss, we model the objective as:

\vspace{-0.2cm}
\begin{equation}\label{eq:gan_loss}
\begin{split}
{L}_{GAN}(u,v)=&log D_V(v)+log (1-D_V(G_U(u))) \\
+&log D_U(u)+log (1-D_U(G_V(v))) \\
\end{split}
\vspace{-0.01cm}
\end{equation}

\noindent where the first row models the adversarial loss for mapping from $U$ to $V$, and the second row defines the adversarial loss for mapping from $V$ to $U$. Specifically,  $G_U(u)$ transforms images of domain $U$ to domain $V$, and $D_V$ aims to classify the
translated image $G_U(u)$ and real image $v$. The adversarial process of $G_U$ aims to minimize the objective against an adversary $D_V$ that tries to maximize it. Similarly, the second row introduces a similar adversarial loss for mapping from $V$ to $U$.

For the cycle consistent reconstruction, $G_U$ and $G_V$ satisfy backward reconstruction consistency as: $u \rightarrow G_U(u) \rightarrow G_V(G_U(u)) \approx u$ and $v \rightarrow G_V(v) \rightarrow G_U(G_V(v)) \approx v$.  Typically, the reconstruction
loss is defined as a pixel wise reconstruction error with L1 or L2 loss~\cite{isola2017image}. Without loss of generality, similar as CycleGAN, we use the L1 reconstruction loss as:

\vspace{-0.5cm}
\begin{equation}\label{eq:rec_loss}
\begin{split}
{L}_R(u,v)=&\vartheta_u\parallel u-G_V(G_U(u))\parallel_1+\\
&\vartheta_v\parallel v-G_U(G_V(v))\parallel_1,
\end{split}
\end{equation}
where $\vartheta_u$,  and $\vartheta_v$ are typically set to 10.

Given the detailed adversarial loss in Eq.\eqref{eq:gan_loss} and the reconstruction loss in Eq.\eqref{eq:rec_loss},  CycleGAN aims to solve the following objective function:

\vspace{-0.1cm}
\begin{equation}\label{eq:cyclegan_aim}
\begin{split}
G^*_U,G^*_V,D^*_U,D^*_V =\arg \min_{G_U,G_V} \max_{D_U,D_V}\mathbb{E}_{u,v\sim p_{data}}L(u,v).
\end{split}
\end{equation}

 \section{The Proposed Model}
In this section, we propose an overall \emph{Q}uality-aware \emph{GAN}~(QGAN) framework for unpaired image-to-image translation, which is based on CycleGAN. As shown in the overall loss function~(Eq.\eqref{eq:dualgan_loss}) of CycleGAN, it has a GAN-based loss term and a pixel-to-pixel level reconstruction loss.
In fact, as the generated images are finally evalua
ted by human, it is important to generate visually qualified images. Nevertheless, human rely on the  high level abstractions of images for perceptual quality evaluation, which is neglected in this process. To generate quality-aware image translations, it is important to incorporate the objective of QGAN with human visual quality constraints. Thus, we define an overall loss function of the proposed QGAN framework as:

\vspace{-0.15cm}
\begin{equation}\label{eq:QGAN_loss}
\vspace{-0.2cm}
\begin{split}
L(u,v)=&L_{GAN}(u,v)+L_{R}(u,v)+ L_{Q}(u,v)\\
\end{split}
\vspace{-0.2cm}
\end{equation}

\noindent where $L_{R}(u,v)$ and $L_{GAN}(u,v)$ share the similar formulations as the unpaired image translation task as introduced before. $L_{Q}(u,v)$ directly measures the quality of reconstructed images in $U$ ~(i.e, $G_V(G_U(u))$) and real images in $U$~(i.e, $u\in U$), and the  reconstructed images in $V$ ~(i.e, $G_U(G_V(v))$) and real images in $V$~(i.e, $v\in U$).
By comparing the quality between each image and its reconstructed version, the proposed QGAN framework could be generally applied to unpaired image-to-image translation without any paired images.
In the following of this section, we provide two detailed implementations of the QGAN framework, i.e., two methods of the quality-aware loss $L_{Q}(u,v)$. We present the overall ideas of our proposed two models, as well as the CycleGAN model in Fig.~\ref{Fig:quality_awareGANs}. Specifically, we would first show how to implement a detailed quality-aware loss based on the classical image quality assessment measures~(middle part of the Fig.~\ref{Fig:quality_awareGANs}). Besides, as CNNs show a huge success for capturing the higher content structure information, instead of applying pretrained deep networks for quality assessment~\cite{johnson2016perceptual,ledig2017photo,zhang2018unreasonable,guo2018automatic}, we would also like to explore whether it is possible to  adaptively model the high level quality loss ~(right part of the Fig.~\ref{Fig:quality_awareGANs}). Next, we introduce the implementations of these two quality-aware losses in detail.

\subsection{QGAN\_A: Classical IQA Loss}
In Section~\ref{sec:related}, we introduce some classical methods for IQA.
As mentioned before, as most BR based IQA models either needed additional labeled image quality values or were only applicable to a specific domain, these BR based quality measures are not suitable for modeling the quality loss in the QGAN framework. Therefore, we plan to adopt FR based IQA models for quality loss modeling. In fact, there are various computational FR models for IQA by measuring the image quality consistently with human subjective evaluations, such as SSIM~\cite{wang2004image}, FSIM~\cite{zhang2011fsim} and GMSD~\cite{xue2014gradient}. Since the focus of this paper is not to design more sophisticated IQA measures, we choose FSIM method~\cite{zhang2011fsim} for modeling the quality loss $L_{Q}(u,v)$ as it is profound with physiological and psychophysical evidences, and showed great success for modeling IQA. We call this proposed model as QGAN\_A~(Assessment).

Specifically, as well researched by physiologists and neuroscientists, visually discernable features coincide with those points that Fourier waves at different frequencies have congruent phases. Thus, by transforming images into a frequency domain, FSIM selects two low-level features based on the phase consistency and gradient magnitude. Additionally, the color characteristics are added to establish the FSIM. For more details of FSIM, please refer to Zhang et al.~\cite{zhang2011fsim}. Thus, we build the $L_{Q}(u,v)$ based on FSIM as:

\vspace{-0.25cm}
\begin{equation}\label{eq:QGAN_a_loss}
\vspace{-0.1cm}
\begin{split}
L_{Q}(u,v)=&\alpha_u[1-FSIM(u,G_V(G_U(u)))]_1+\\
&\alpha_v[1-FSIM(v,G_U(G_V(v)))]_1,
\end{split}
\vspace{-0.2cm}
\end{equation}

\noindent where $\alpha_u$ and $\alpha_v$ are regularization parameters. $FSIM(x,y)$ is the quality similarity score calculated using the FSIM~\cite{zhang2011fsim} algorithm.
Then, by minimizing the loss function~$L_{Q}(u,v)$, we encourage the generators~$(G_U,G_V)$ to generate images such that each input image and its reconstructed image have similar quality scores. In contrast, if the generators~$(G_U,G_V)$ could not accomplish the translation task well, the $FSIM$ value between  an input image and its reconstructed image is closer to 0. Then, the loss function~$L_{Q}(u,v)$ becomes larger. Therefore, based on FSIM loss, the quality of the generated image can be constantly utilized to induce a positive effect.


\subsection{QGAN\_C: Adaptive Content-based Loss}

In this subsection, we introduce how to design an adaptive perceptual quality-aware loss for unpaired image translation. We call the proposed QGAN framework QGAN\_C~(Content) as it models the higher level content of deep networks.

In fact, some previous works  have shown that high-quality images can be generated by defining the high level perceptual losses based on high level features from pretrained neural networks~\cite{johnson2016perceptual,ledig2017photo}. The intuition is that, the pixel-based loss functions focus on low level image information, which optimize the pixel-wise average between the ground truth images and the generated images and then lead to over-smooth results. Instead, the perceptual losses optimize the high level content losses from pretrained VGG networks, encouraging the ground truth images and the generated images are similar in the VGG feature space.

Despite the visually superior performance, we argue that these previous works have some limitations. Specifically, these pretrained VGG networks are capable of extracting high level features that are well trained for specific classification tasks.  The GAN-based approach also naturally achieves the high level feature learning of images with an adversarial learning process. In fact, as the high level features extracted from pretrained VGG networks are optimized for the image classification task, they are inferior when transferred to  image translation task. Thus, we argue that it is better to design an adaptive content-based loss from the high level features from the GAN-based approach that is tailored to the image translation task. Next, we would detail how to design the adaptive content-based loss directly from the GAN framework.

Specifically, instead on a pretrained VGG network in related works~\cite{johnson2016perceptual,ledig2017photo}, we define the content-based loss as:

\vspace{-0.1cm}
\begin{align}\label{eq:QGAN_c_loss}
\vspace{-0.1cm}
{L}_{Q}(u,v)=&\beta_u\parallel \phi_{i}(u)-\phi_{i}(G_V(G_U(u)))\parallel_1\nonumber+\\
&\beta_v\parallel \phi_{i}(v)-\phi_{i}(G_U(G_V(v)))\parallel_1,
\vspace{-0.1cm}
\end{align}

\noindent where $\beta_u$ and $\beta_v$ are regularization parameters, $\phi_{i}$ indicates the feature map located in the i-th convolutional~(after activation) layer of the generator. In practice, we use the popular ``U-NET'' as the generator~\cite{isola2017image}, which is shown in the rightmost part of Fig~\ref{Fig:quality_awareGANs}. As the parameters of the ``U-NET'' change in the adversarial training process, for any image $u$, its content features $\phi_{i}(u)$ adaptively updates during the training process of image translation task.


Please note that,  recently researchers also proposed a Perceptual Adversarial Network~(PAN) for enhancing image-to-image translation quality on the labeled image pairs under the GAN framework~\cite{wang2018perceptual}. In PAN, besides the generative adversarial loss widely used in GANs, a perceptual adversarial loss~(\emph{PA loss}) is introduced to undergo an adversarial training process between the image generation network and the hidden layers of the discriminative network. In designing the~\emph{PA loss}, it is required to compare the quality of each generated image and its corresponding ground truth. PAN also demonstrates the possibility to be extended for the unpaired image  translation task, which is achieved by calculating the difference of the mean features on two domains. Nevertheless, the uniqueness of each individual image is neglected and smoothed by the average operation. Therefore, PAN could not well tackle the unpaired image translation. In contrast, we focus on the unpaired image translation task and  propose an adaptive content-based loss, which considers each individual image  by measuring image quality between each input and its corresponding reconstructed image. In summary, our proposed model is more robust for unpaired image-to-image translation.

\section{Model Training}
Our proposed QGAN framework~(Eq.\eqref{eq:QGAN_loss}) with two detailed quality loss implementations, i.e., QGAN\_A and QGAN\_C, could be trained under a unified optimization framework. We show the training process of QGAN in Alg. 1. It includes training the discriminators $D_A$ and $D_B$~(from 4th to 9th line) and training the generators $G_A$ and $G_B$~(from 11th to 14th line).
In practice, we set the number of iterations~(i.e., $n$)  to be 2-5~\cite{arjovsky2017wasserstein,yi2017dualgan,gulrajani2017improved}. We use the RMSProp \cite{hinton2012rmsprop} solver  as it performs well on highly non stationary problems. We initialize the learning rate for RMSProp as 0.00005. The clip parameter $c$ clips value to a specified range, and $c$ is set in $[0.01,0.1]$~\cite{arjovsky2017wasserstein}. For all experiments, the values of $\alpha_u$ and $\alpha_v$ are set  as 15 times as large as that of $\vartheta_u$  and $\vartheta_v$, $\beta_u$ and $\beta_v$ are set to 20 times as large as that of $\vartheta_u$  and $\vartheta_v$. The feature map in Eq.\eqref{eq:QGAN_c_loss} is chosen as $i=6$, i.e., we choose $\phi_{i}(u)$ as the sixth convolutional~(after activation) layer of the generator for the content-based loss. As the original loss of $L_{GAN}$ is unstable during the training process, similar as many works~\cite{arjovsky2017wasserstein,zhu2017unpaired,yi2017dualgan}, we train~$G_U$ and $G_V$ to minimize~{\small$L_{g}(u,v)\!=\!(D_U(G_V(v))-1)^2+(D_V(G_U(u)-1))^2$\small}, and train~$D_U$ and $D_V$ using~{\small$L_{d}(u,v)\!=\!(1-D_U(u))^2+{D_U(G_V(v))}^2+{(1-D_V(v))}^2+{D_V(G_U(u))}^2$}. Specifically, Alg. 1 shows the training procedure for optimizing the proposed loss function.

For the generators~($G_U$ and $G_V$) and discriminators~($D_U$ and $D_V$), we use the architectures that are widely adopted in image-to-image translation. Specifically, we adopt the ``U-NET'' structure~\cite{isola2017image} for our generative networks as this structure is successfully applied in many image generation taks~\cite{zhu2017unpaired,yi2017dualgan,isola2017image}. The ``U-NET'' architecture contains two stride-2 convolutions, several residual blocks and two fractionally strided convolutions. This network contains 16 layers with skip connection between each layer~$i$ and layer~$16-i$, where~$0<i<9$. Thus, the ``U-NET'' allows information to short across the network, and models more high level structure information in one layer. For the discriminator, we use the same CNN architecture as in~\cite{zhu2017unpaired,yi2017dualgan,isola2017image}, where each discriminator contains four layers.


\vspace{-0.2cm}
\begin{algorithm}[h]
\caption{The Algorithm of QGAN Framework.}
{\bf Input:}
real data $U$, real data $V$, batch size $m$, the number of discriminator iterations per generator iteration $n$, generator parameters~$\Theta_U$ and $\Theta_V$, discriminator parameters $\Omega_U$ and $\Omega_V$, clipping parameter~$c$.
\begin{algorithmic}[1]
\State Randomly initialize $\Theta_i$, $\Omega_i$, $i \in\{U,V\}$
\Repeat
 \For{$t=1$ to $n$}
   \State get mini-batch $\{u^{(i)}\}^m_{i=1}$ from the real data $U$
   \State get mini-batch $\{v^{(i)}\}^m_{i=1}$ from the real data $V$
   \State $d \leftarrow \tfrac{1}{m}\sum_{i=1}^{m}L_d(u^{(i)},v^{(i)})$
      \State update~$\Omega_U, \Omega_V \leftarrow$ RMSProp optimizer $d$
   \State clip($\Omega_U$, $-c$, $c$) \{Clip the weight of $D_U$ \}
      \State clip($\Omega_V$, $-c$, $c$) \{Clip the weight of $D_V$ \}
 \EndFor
       \State get mini-batch $\{u^{(i)}\}^m_{i=1}$ from the real data $U$
   \State get mini-batch $\{v^{(i)}\}^m_{i=1}$ from the real data $V$
   \State {\small$g\leftarrow \tfrac{1}{m}\sum_{i=1}^{m}[L_{G}(u^{(i)},v^{(i)})\!+\!L_R(u^{(i)},v^{(i)})\!+\!L_Q(u^{(i)},v^{(i)})]$}
    \State update $\Theta_U, \Theta_V \leftarrow$ RMSProp optimizer $g$
\Until convergence
\end{algorithmic}
\end{algorithm}

\section{Experiments}


In this section, we  conduct extensive experimental results to show the  quality improvement of our proposed two methods~(i.e., {\small QGAN\_A} and {\small QGAN\_C}) on unpaired image-to-image translation.
We perform experiments on four datasets that are widely used for image translation: {\small PHOTO-SKETCH}~\cite{yi2017dualgan}, {\small LABEL-FACADE}~\cite{yi2017dualgan}, {\small OIL-CHINESE}~\cite{yi2017dualgan}, {\small SUMMER-WINTER}~\cite{zhu2017unpaired}. Table~\ref{tab:freq} shows the details of all these datasets, where the training and testing images are automatically divided in these datasets. The number of training images contains the images from both domains for training. Specifically, the {\small PHOTO-SKETCH} and the {\small LABEL-FACADE} datasets include the paired images between the corresponding two domains. In the model training process, we omit the paired correspondence on these two datasets, and we use the ground truth~(GT) of the pair relationships for better visual evaluation.

For fair comparisons, firstly we choose CycleGAN~\cite{zhu2017unpaired} as a baseline as it is a state-of-the-art unpaired image translation model, and it shares similar ideas with many unpaired image-to-image translation models, including DualGAN~\cite{yi2017dualgan} and DiscoGAN~\cite{kim2017learning}. In the experimental setup process, we use the same settings as CycleGAN. Besides, as the pretrained VGG based perceptual loss functions have been widely used for enhancing the quality of the image translation tasks~\cite{ledig2017photo,johnson2016perceptual}, we combine the metrics of the pretrained VGG based perceptual loss function with the optimization function of CycleGAN as a baseline. We call this baseline as  CycleGAN-VGG. The comparison between QGAN\_C and CycleGAN-VGG would show whether the adaptive content loss in QGAN\_C is superior than the pretrained perceptual loss from the VGG.

\vspace{-0.2cm}
\begin{table}[H]
\vspace{-0.1cm}
  \caption{Dataset Description.}
  \label{tab:freq}
  \vspace{-0.3cm}
  \begin{tabular}{ccc}
    \toprule
    Dataset  & \tabincell{c}{Num. of training images}  & \tabincell{c}{Num. of test images}\\
    \midrule
    PHOTO-SKETCH & 1990 & 388 \\
    OIL-CHINESE & 2354  & 94 \\
    SUMMER-WINTER & 1924 & 476 \\
    LABEL-FACADE & 800  & 200 \\
   \bottomrule
\end{tabular}
\vspace{-0.3cm}
\end{table}


\subsection{Experimental Results}
\begin{figure}
\begin{center}
\includegraphics[width=3in]{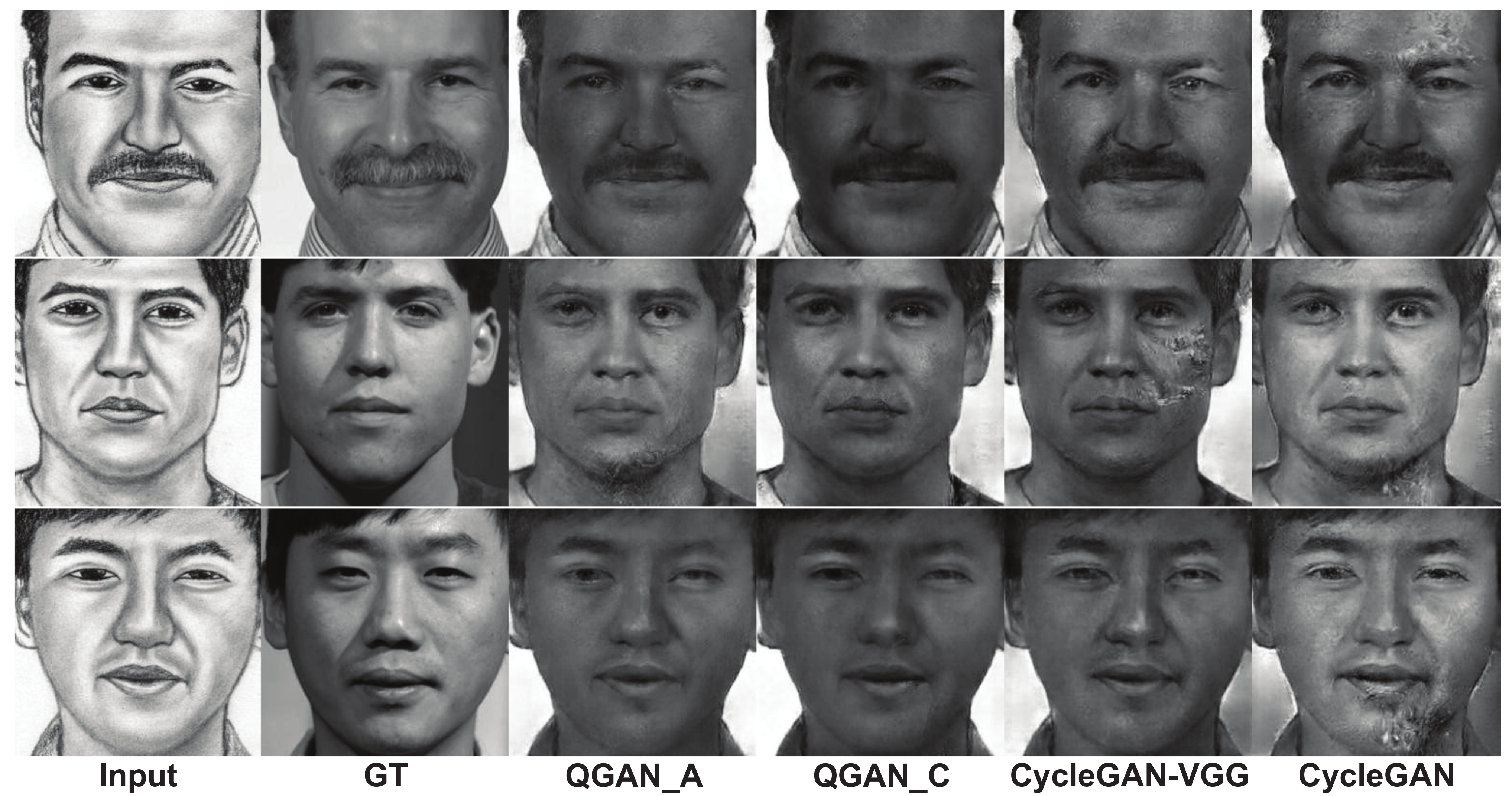}
\vspace{-0.3cm}
\caption{Results of sketch$\rightarrow$photo translation.}
\label{Fig:sketch2photo2}
\end{center}
\vspace{-0.8cm}
\end{figure}

\begin{figure}
\begin{center}
\includegraphics[width=3in]{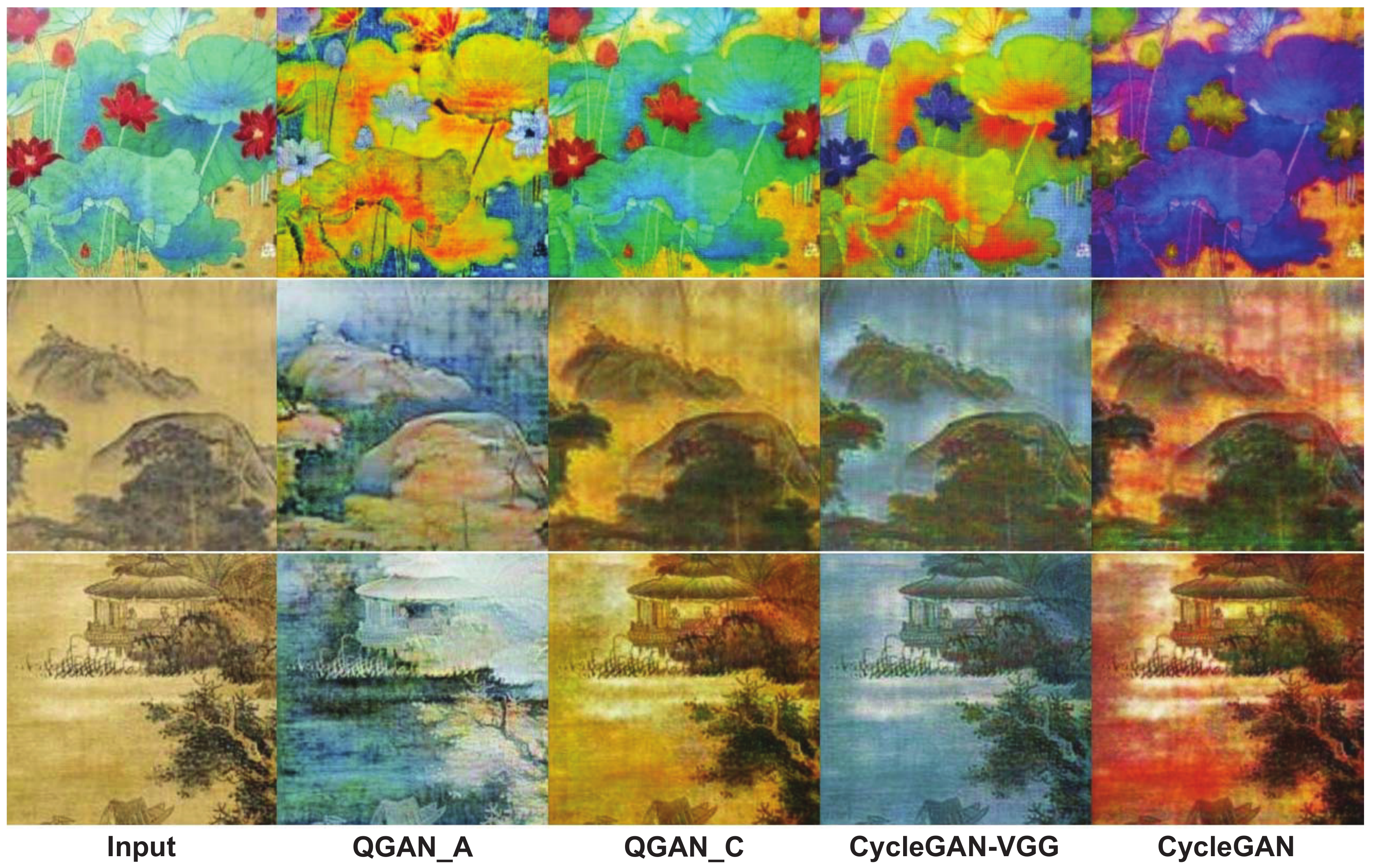}
\vspace{-0.3cm}
\caption{Results of Chinese$\rightarrow$oil translation.}
\label{Fig:chinese2oil}
\end{center}
\vspace{-0.5cm}
\end{figure}

\begin{figure}
\begin{center}
\includegraphics[width=3in]{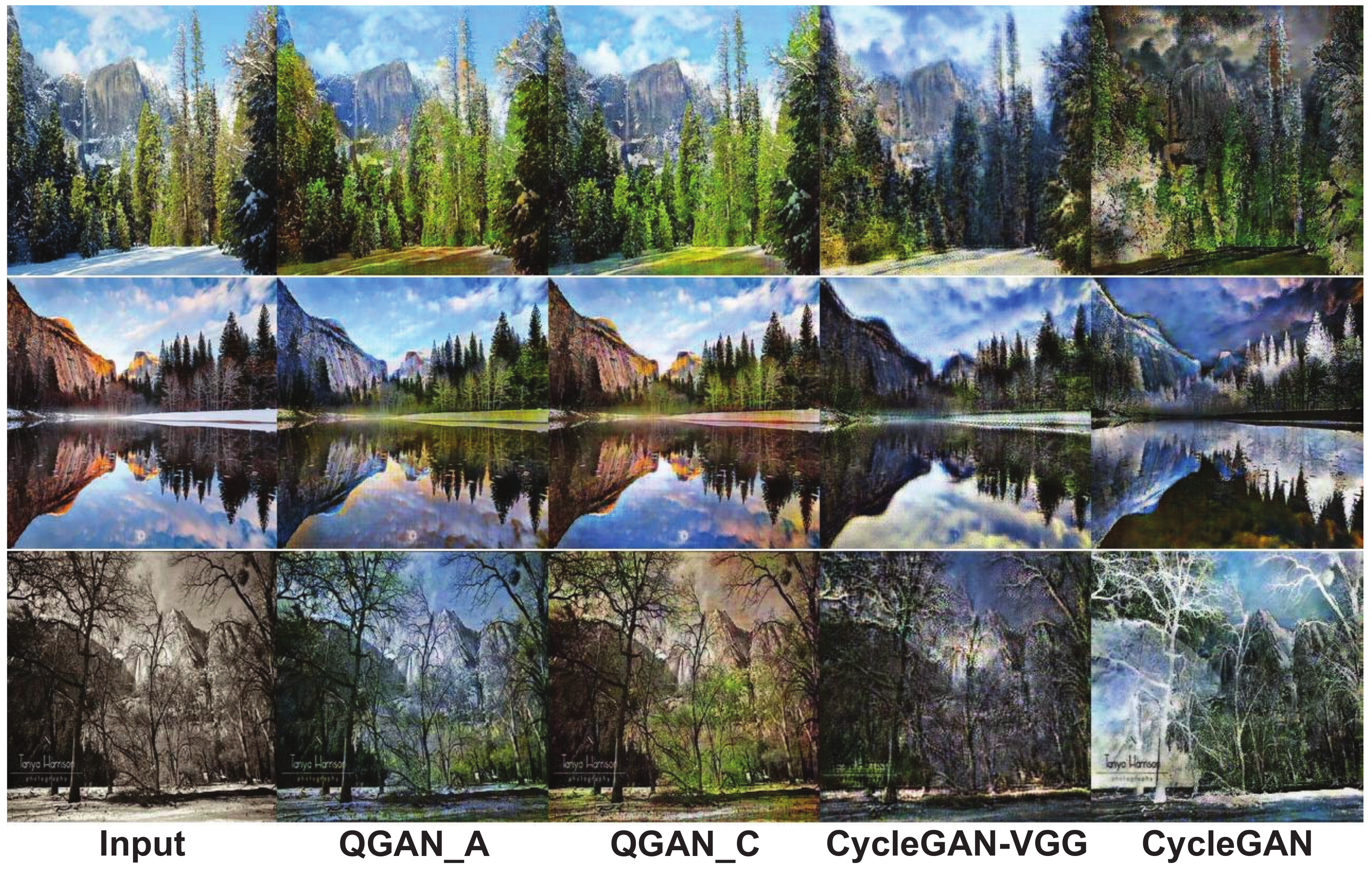}
\vspace{-0.3cm}
\caption{Results of winter$\rightarrow$summer translation.}
\label{Fig:winter2summer}
\end{center}
\vspace{-0.5cm}
\end{figure}

\begin{figure}
\begin{center}
\includegraphics[width=3in]{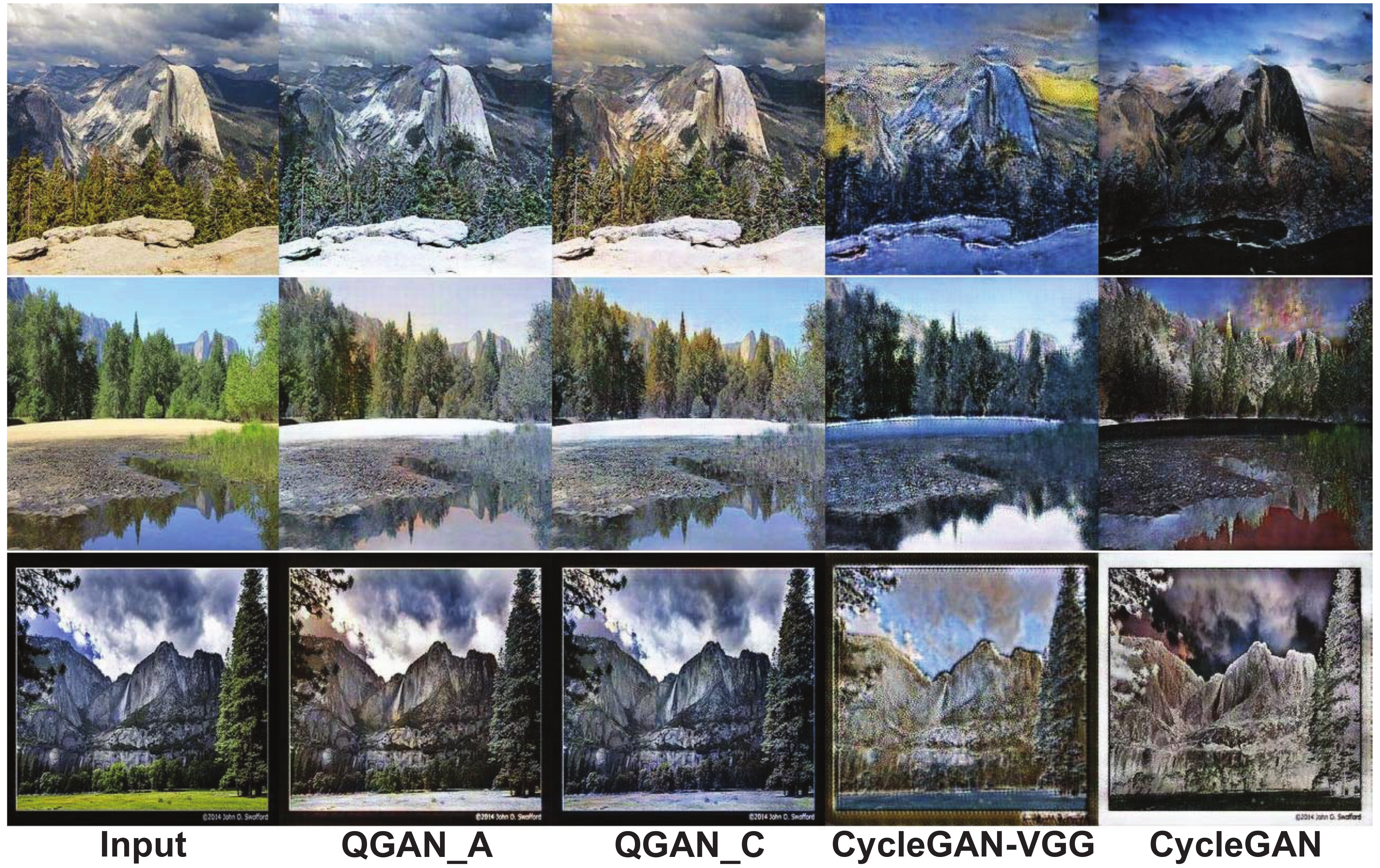}
\vspace{-0.3cm}
\caption{Results of summer$\rightarrow$winter translation.}
\label{Fig:summer2winter}
\end{center}
\vspace{-0.5cm}
\end{figure}

\begin{figure}
\begin{center}
\includegraphics[width=3in]{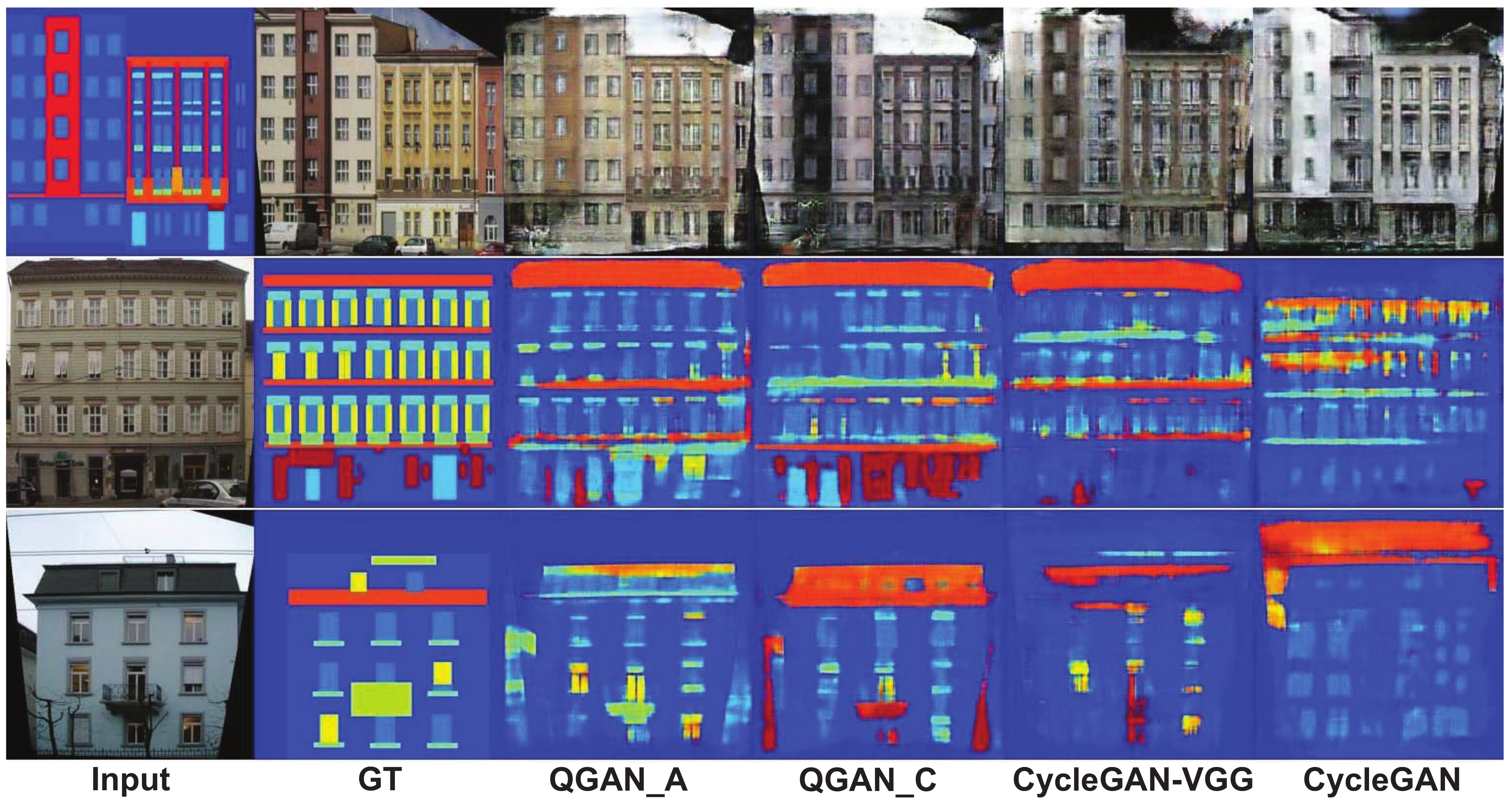}
\vspace{-0.3cm}
\caption{Results of label$\leftrightarrow$ facade translation.} 
\label{Fig:label2facade}
\end{center}
\vspace{-0.5cm}
\end{figure}

We perform extensive experimental results on four widely used tasks: sketch$\rightarrow$photo (Fig.~\ref{Fig:sketch2photo2}), Chinese$\rightarrow$oil (Fig.~\ref{Fig:chinese2oil}),  summer$\leftrightarrow$winter (Fig.~\ref{Fig:sketch2photo1},~\ref{Fig:winter2summer},~\ref{Fig:summer2winter}),  and label$\leftrightarrow$ facade (Fig.~\ref{Fig:label2facade}). In all these tasks, the \emph{Input} shows the input images from the source domain, and \emph{QGAN\_A} and \emph{QGAN\_C} are our proposed two methods.  For better comparison, we also show the ground truth if the dataset contains the ground truth of the paired images. As can be seen from these figures, in almost all tasks, compared to CycleGAN and CycleGAN-VGG, our proposed two models generate quality-enhanced images with less blurry parts. E.g., in Fig.~\ref{Fig:sketch2photo2}, the results of QGAN are less blurry with more details, while there exists blurry parts on the forehead that appeared in CycleGAN and CycleGAN-VGG.


We observe a typical phenomenon in Fig.~\ref{Fig:winter2summer} and Fig.~\ref{Fig:summer2winter}, which performs image translation in domains of winter and summer. A key characteristic of the domain translation task is that snow would fall on objects in winter, while it disappears in summer. E.g., as shown in the upper-right corner in  the last column of Fig.~\ref{Fig:summer2winter},  the result of CycleGAN shows colorful patches.
To analyze this phenomenon, we find that as the snow falls on the objects in an image taken in winter, some parts in this image contain colorful patches~(e.g., the green mountain with white snow). The generator in CycleGAN would indecisively pick a color~(e.g., the color of the snow or the color of the mountain) as this decision would not have a large impact on the pixel-wise based reconstruction loss. Luckily, our proposed two models could automatically distinguish the varying frequency of these colorful patches. 
Besides, we observe that snow is also not well captured in CycleGAN-VGG. We guess a possible reason is that CycleGAN-VGG relied on the pretrained network parameters from the ImageNet dataset, which is designed on the classification task with specific categories.  As snow usually appears in limited images with small regions, it is also not well captured by the VGG network. In our proposed two models, the adaptive quality-aware loss is optimized for the current image translation domain, in order to avoid the deficiencies of using pretrained network based perceptual loss. To better show the effectiveness of using the adaptive content features~(\emph{QGAN\_C}) compared to the pretrained network, we present the results of the label$\leftrightarrow$facade task as an example. The pretrained ImageNet dataset does not contain the label category. As shown in Fig.~\ref{Fig:label2facade},  it is obvious that \emph{QGAN\_C} generate  more details of edge. E.g., as shown in the last column,  the outline of the label in  \emph{CycleGAN-VGG} disappears. In contrast, \emph{QGAN\_C} generates more detailed label information compared to the baselines.  

When comparing our proposed two quality losses, the results of QGAN\_C look similar to the results of QGAN\_A, but QGAN\_A sometimes produces images with better quality that meet human perception, such as in Fig.~\ref{Fig:sketch2photo2}. We leave the quantitative quality comparisons of these two proposed models in the next part. To analyze the main reason, the adaptive content loss in QGAN\_C  relies on the training of the generator and it is more likely to be affected by the instability of the training process.  In contrast, the classical IQA based loss function in QGAN\_A is stable and independent of the process of GAN-based training, and it is easier in model tuning process.

\subsection{Quantitative Evaluation}
To better compare the quality of the generated images from different models, in this part, we conduct quantitative evaluation under different quality metrics. Specifically, we adopt three commonly used full reference image quality assessment methods: SSIM~\cite{wang2004image}, FSIM~\cite{zhang2011fsim} and GMSD~\cite{xue2014gradient}. These three methods measure the image quality that reflect the human visual experience from various aspects. The larger values of SSIM and FSIM denote better quality, and the smaller scores of GMSD denote better quality. Since all these measures rely on the paired image information, for the two tasks of sketch$\rightarrow$photo and label$\rightarrow$facade  with ground truth of paired information, we can calculate the quality measure with the generated image and the ground truth in the test evaluation process. With the unpaired images in the remaining two datasets, we could not calculate these measures. It can be seen from Table~\ref{tab:oneQE} that our proposed methods perform consistently better than CycleGAN-VGG and CycleGAN under the three measures. E.g., QGAN\_A improves over CycleGAN-VGG about 5\%  and more than 8\% improvement over CycleGAN under the SSIM measure in sketch$\rightarrow$photo task. Since these metrics evaluate human visual experience from various perspectives, we could empirically conclude the effectiveness of quality enhancement of our proposed models. Last but not least, by comparing the results of QGAN\_A and QGAN\_C, we observe that for each assessment measure, the results of QGAN\_A always outperform QGAN\_C.


\begin{table}[h]
\centering
\vspace{-0.3cm}
\caption{Quality score evaluated on sketch$\rightarrow$photo and label$\rightarrow$facade with different methods. We calculate each measure  between the generated image and its corresponding Ground Truth~(GT).}\label{tab:oneQE}
\vspace{-0.25cm}
\begin{scriptsize}
\begin{tabular}{p{1.1cm}cccp{1.1cm}<{\centering}c}
\hline
\multicolumn{2}{c}{Input and Method}
& { QGAN\_A}
& { QGAN\_C}
& { CycleGAN-VGG}
& { CycleGAN} \\\hline
\multirow{3}*{\tabincell{c}{GT and\\ sketch$\rightarrow$photo}}
& SSIM & \textbf{0.5683} & 0.5681 &0.5412 & 0.5256 \\
& FSIM & \textbf{0.7749} & 0.7746 &0.7658 & 0.7616 \\
& GMSD & \textbf{0.2128} & 0.2145 &0.2148 & 0.2146 \\\hline
\multirow{3}*{\tabincell{c}{GT and\\ label$\rightarrow$facade}}
& SSIM &\textbf{ 0.1686} & 0.1233 &0.0710 & 0.0545 \\
& FSIM & \textbf{0.6058} & 0.588  & 0.5897 & 0.5719 \\
& GMSD &\textbf{ 0.2816} &0.2855  &0.2849 & 0.2901 \\\hline
\end{tabular}
\vspace{-0.5cm}
\end{scriptsize}
\end{table}


\subsection{Mean Opinion Score (MOS) Testing}
Assessing the quality of generated image is an open question. Though we have conducted various measures for image quality assessment in previous part, human visual experience is the golden standard for assessing the quality of generated images. Thus, to better compare our proposed models with the baselines, we conduct a \emph{M}ean \emph{O}pinion \emph{S}core~(MOS) testing by human evaluation. This MOS testing avoids the shortcomings of each quality evaluation metric and gives the overall perceptual experience. To realize this, we design a MOS system that asks each rater to give a numerical indication of the perceived quality of each generated image from each method. Specifically, we ask 24 raters to assign a score from 1-5, where 1 denotes the lowest perceived quality, and 5 is the highest perceived quality. In the system design process, the images generated by different methods are listed randomly. Also, we randomly repeat some images in the system to see whether the rater gives the same rating to the same image that appears in different orders. We remove the raters that give different ratings to the same image. We evaluate on the  four tasks: sketch$\rightarrow$photo, Chinese$\rightarrow$oil, label$\rightarrow$facade, and winter$\rightarrow$summer. Since manual scoring is time-consuming and expensive, researchers often run small datasets with a random selection to approximate human perception, we randomly select a third of all test image of four tasks. Thus, each rater rated four tasks with 772 images. The final MOS scores of all methods are summarized in Table~\ref{tab:mos}. As can be seen from this table, the MOS testing results show that QGAN framework outperforms CycleGAN and CycleGAN-VGG on all tasks.
Generally, CycleGAN-VGG shows better results compared to CycleGAN by adding the pretrained perceptual loss, and our proposed QGAN\_C further improves CycleGAN-VGG with the adaptive perceptual loss. When comparing the results of all models, generally our proposed QGAN\_A shows the best performance, followed by our proposed QGAN\_C model. However, we observe that for the Chinese$\rightarrow$oil translation task, QGAN\_C shows better results than QGAN\_A with MOS testing. Meanwhile, in Table~\ref{tab:oneQE}, the quantitative results show that  QGAN\_A performs better than QGAN\_C on four quality measures. This inconsistency between the quality measures and the MOS score also observed by previous works~\cite{ledig2017photo}. We guess a possible reason  is that human visual evaluation is quite subjective, and each quality measure could only partially reflect human visual experience. Nevertheless, as our proposed two models consistently outperform the baselines to a large margin, we could conclude the effectiveness of our proposed two models for quality-enhanced unpaired image translation. Also, our proposed QGAN\_A that relies on the classical image quality measures outperforms QGAN\_C in most situations.

\begin{table}[h]
\centering
\vspace{-0.3cm}
\caption{MOS score for different methods on all tasks.}\label{tab:mos}
\vspace{-0.3cm}
\begin{scriptsize}
\begin{tabular}{cccccc}
\hline
\multicolumn{1}{c}{Tasks}
& {QGAN\_A}
& {QGAN\_C}
& { \tabincell{c}{CycleGAN-VGG}}
& { \tabincell{c}{CycleGAN}}  \\\hline
\multirow{1}*{\tabincell{c}{sketch$\rightarrow$photo}}
& \textbf{2.71} & 2.63  &2.65& 2.55 \\\hline
\multirow{1}*{\tabincell{c}{label$\rightarrow$facade}}
& \textbf{3.26} & 3.03  &2.91& 2.61 \\\hline
\multirow{1}*{\tabincell{c}{Chinese$\rightarrow$oil}}
& 3.13 &\textbf{3.73}   &2.95&2.79\\\hline
\multirow{1}*{\tabincell{c}{winter$\rightarrow$summer}}
& \textbf{3.91} & 3.76  &2.56& 2.06 \\\hline
\end{tabular}
\end{scriptsize}
\vspace{-0.5cm}
\end{table}

\subsection{Analysis of The Objective Function}
In our proposed QGAN framework, in addition to the adversarial loss, there are two important regularization terms: a reconstruction loss $L_R(u,v)$ and a quality loss $L_Q(u,v)$. In this part, we would demonstrate the effectiveness of the proposed framework from these two aspects.

%

\textbf{Impact on the reconstruction loss.} To study the effect of the reconstruction loss in QGAN framework, we design a simplified  model that discards the reconstruction loss in QGAN  framework and redefine the objective function as:

\begin{equation}\label{eq:testOne}
\vspace{-0.17cm}
\begin{split}
L(u,v)=&L_{GAN}(u,v)+L_{Q}(u,v).
\end{split}
\vspace{-0.17cm}
\end{equation}

\begin{figure}
\begin{center}
\includegraphics[width=3in]{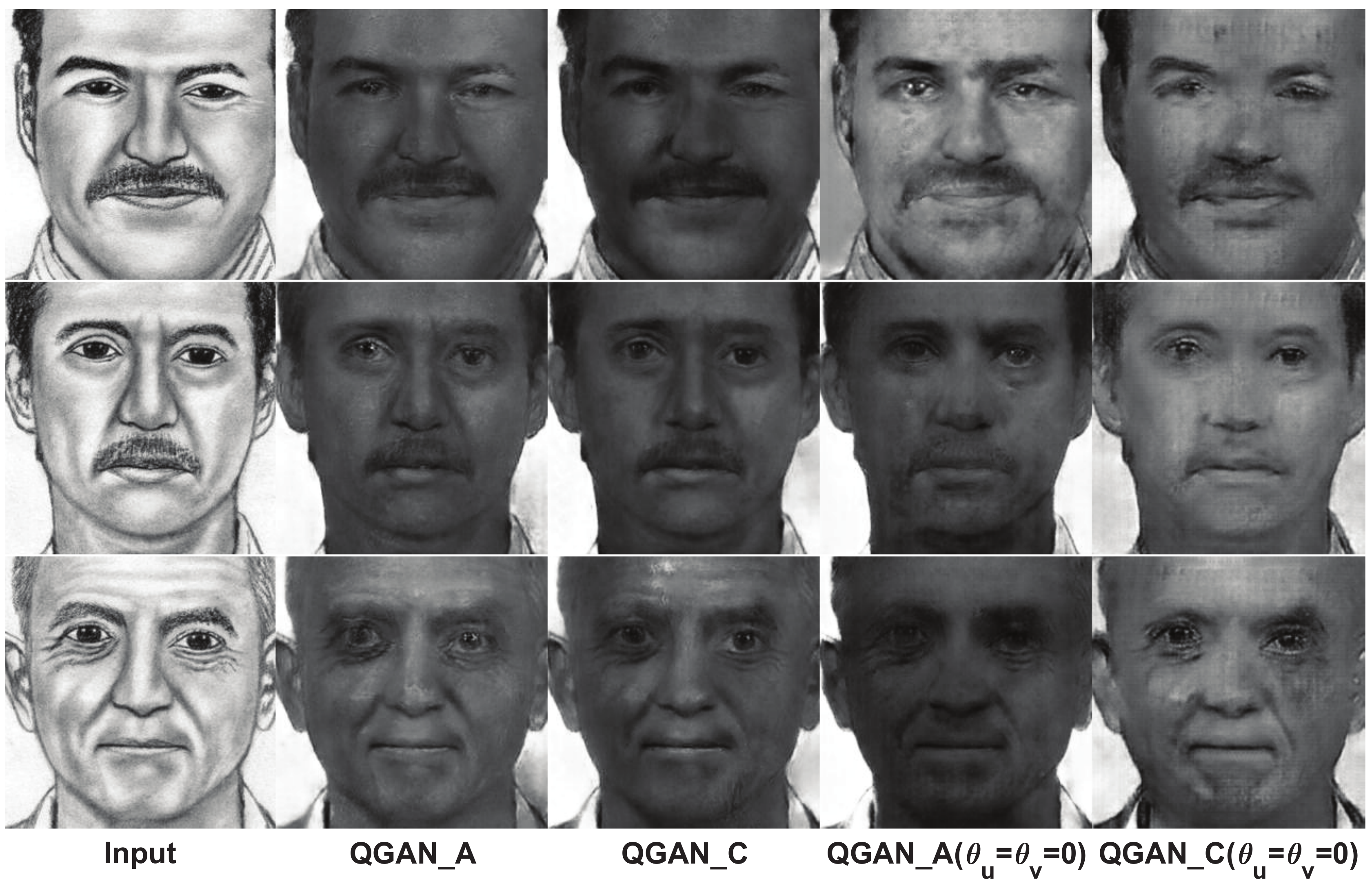}
\vspace{-0.2cm}
\caption{Variants of our proposed methods for the task of sketch$\rightarrow$photo. Specifically, the last two columns are two simplifications of our proposed models without the reconstruction loss.}
\label{Fig:onlyfsim}
\end{center}
\vspace{-0.5cm}
\end{figure}

In other words, the regularization terms in Eq.\eqref{eq:rec_loss} of the reconstruction loss are set as: $\theta_u\!=\!\theta_v\!=\!0$. We show some qualitative examples in Fig.~\ref{Fig:onlyfsim} under this setting, where the last two columns are the generated images without the reconstruction loss in our proposed two models, respectively. We observe that the generated images are blacker in color with more noisy points as compared to the QGAN framework that considers the reconstruction loss.   We guess a possible reason is that  QGAN\_A relies on the FSIM based quality loss that emphasizes on the low-level features in frequency. Thus, only relying on quality assessment based loss without any reconstruction loss makes it unstable in the training process, leading to a smooth distribution that prefers black images. QGAN\_C($\theta_u\!=\!\theta_v\!=\!0$) relies on the higher level content structure based similarity without any pixel level constraint, thus also degrades quality of the generated images. Therefore, we conclude that all terms are critical in the model training process.

\textbf{Impact on the choice of the classical image quality measures.} Image quality assessment models could be classified into FR and BR measures. In our proposed QGAN\_A model, we use the FR based quality measure for modeling the quality loss. Thus, it is natural to ask, is it possible to define image quality loss based on the BR methods? To answer this question, in this part we evaluate the effects of using a blind reference IQA measure based loss in the proposed QGAN framework. Without loss of generality, we select NIQE~\cite{mittal2013making} as a typical blind-reference method to evaluate the image quality. NIQE  uses measurable deviations from statistical regularities observed in natural images without labeling on human-rated distorted images. Hence, it is suitable for many image-to-image translation tasks without any human labeling effort. A smaller score of NIQE indicates better perceptual quality, so we change the classical IQA loss function in Eq.\eqref{eq:QGAN_a_loss} and redefine it as:

\vspace{-0.4cm}
\begin{small}
\begin{equation} \label{eq:QGAN_br}
\begin{split}
{L}_Q(u,v)=&\vartheta_u NIQE(G_V(G_U(u)))+\\
&\vartheta_v NIQE(G_U(G_V(v))),
\end{split}
\vspace{-0.17cm}
\end{equation}
\end{small}

\noindent where $NIQE(x)$ calculates the no-reference image quality score for image $x$ using the NIQE method.

\begin{figure}
\begin{center}
\includegraphics[width=3in]{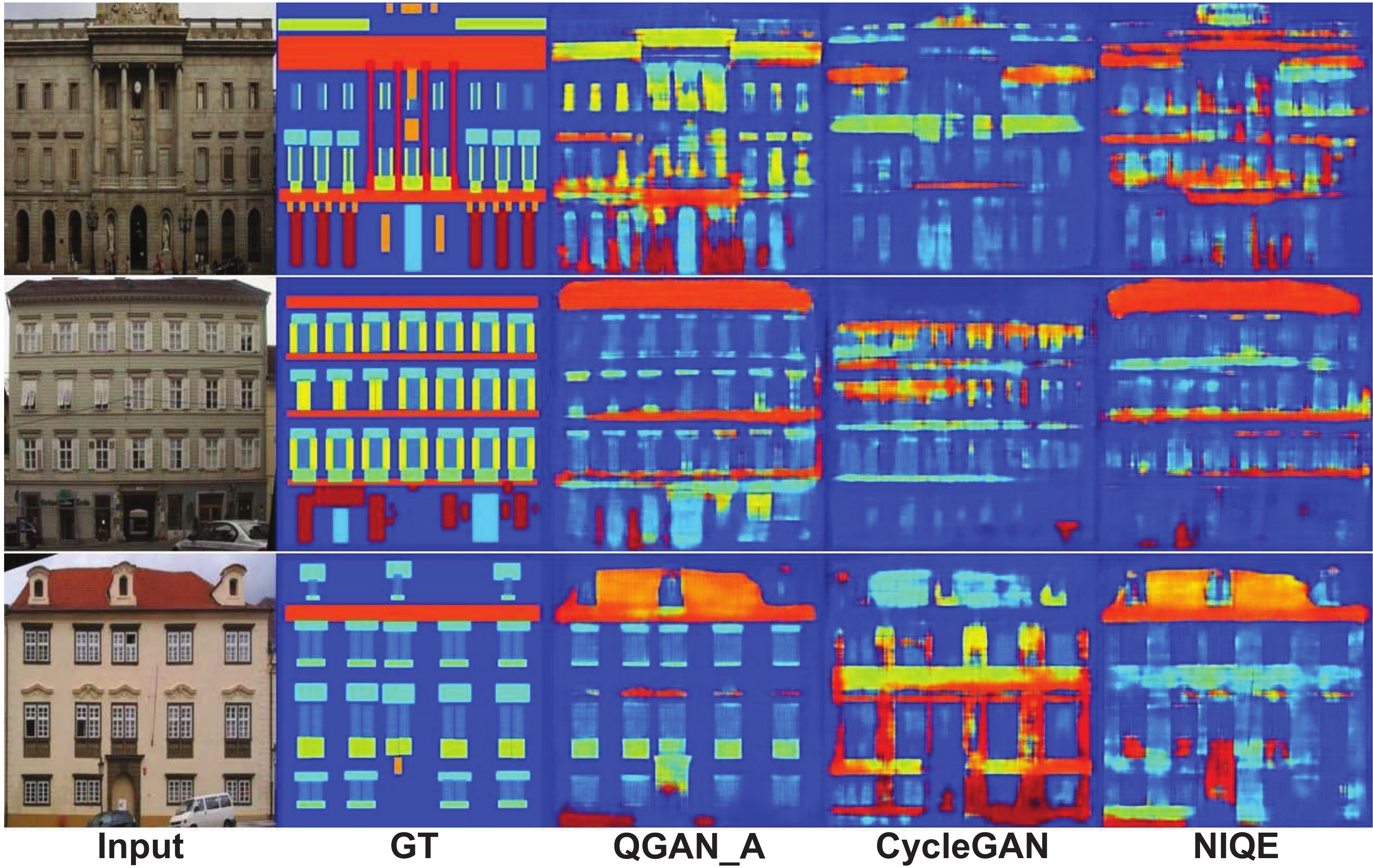}
\vspace{-0.3cm}
\caption{Variants of our proposed methods for the task of facade$\rightarrow$label.}
\label{Fig:niqe}
\end{center}
\vspace{-0.7cm}
\end{figure}

In Fig.~\ref{Fig:niqe}, we show the facade$\rightarrow$label task of the QGAN results with the above defined BR quality loss, where the last column shows the results of using NIQE based loss in the QGAN framework. It is visually obvious that the NIQE based quality loss does not improve the quality of generated images compared to CycleGAN. In fact, despite NIQE can compute the quality for an image, NIQE is based on a certain priori knowledge about natural images. For different tasks, the target images are possibly  not consistent with the certain priori knowledge. If we blindly minimize the score of NIQE, the generated images are hard to meet human visual experience. For example, in label$\rightarrow$facade task~\cite{yi2017dualgan}, the NIQE score of a label image is about 6, while the NIQE score of a facade image is about 9. Please note that, besides the NIQE measure, nearly all blind reference based models need to rely on the prior assumption of the image domains. Thus, we conclude that FR based measures are more suitable in designing image quality-aware loss evaluation.

\subsection{Discussion of Our Proposed Two Models}
In our proposed QGAN framework, there are two kinds of models~(QGAN\_A and QGAN\_C). In this subsection, we compare the  experimental results and discuss the strengths and weaknesses of them.

QGAN\_A relies on a static classical IQA measure, i.e., FSIM, to calculate the quality loss. As FSIM models the phase congruency and other features that are based on physiological and psychophysical evidences from human visual systems, the FSIM based quality loss is intuitive to understand. Besides, this classical IQA based measure does not rely on any intermediate architecture of the generator. Therefore, QGAN\_A does not introduce additional parameters in the model training process and is empirically easier to train compared to QGAN\_C. However, there are also limitations of of QGAN\_A. On one hand, as there is no universal golden-standard metric for image quality evaluation, the performance is limited by the specific IQA based quality measure. E.g., the  phase congruency (PC) feature in FSIM is sensitive to image noise\cite{kovesi2000phase}. On the other hand, as the calculation of FSIM is computationally expensive, the runtime of QGAN\_A is longer than QGAN\_C.

Different from QGAN\_A, QGAN\_C does not introduce any prior knowledge of IQA and relies on an adaptive content-based loss. Specifically, QGAN\_C uses the features extracted by the generators, so it can deal with any dataset. Furthermore,  the ``U-NET" architecture of generators can eliminate noisy effects from inputs. By comparing the feature maps extracted from this architecture, denoising effect is automatically achieved for some noisy input images. However, as QGAN\_C is an adaptive algorithm, and the proposed adaptive content loss relies on the intermediate feature extractor of the generator, this proposed model introduces additional parameters, i.e., feature map $\phi_i$ after the $i$-th convolutional layer. In practice, we find this parameter adds additional tuning complexity considering the
instability in the GAN training process. 

\begin{figure}
\begin{center}
\includegraphics[width=3in]{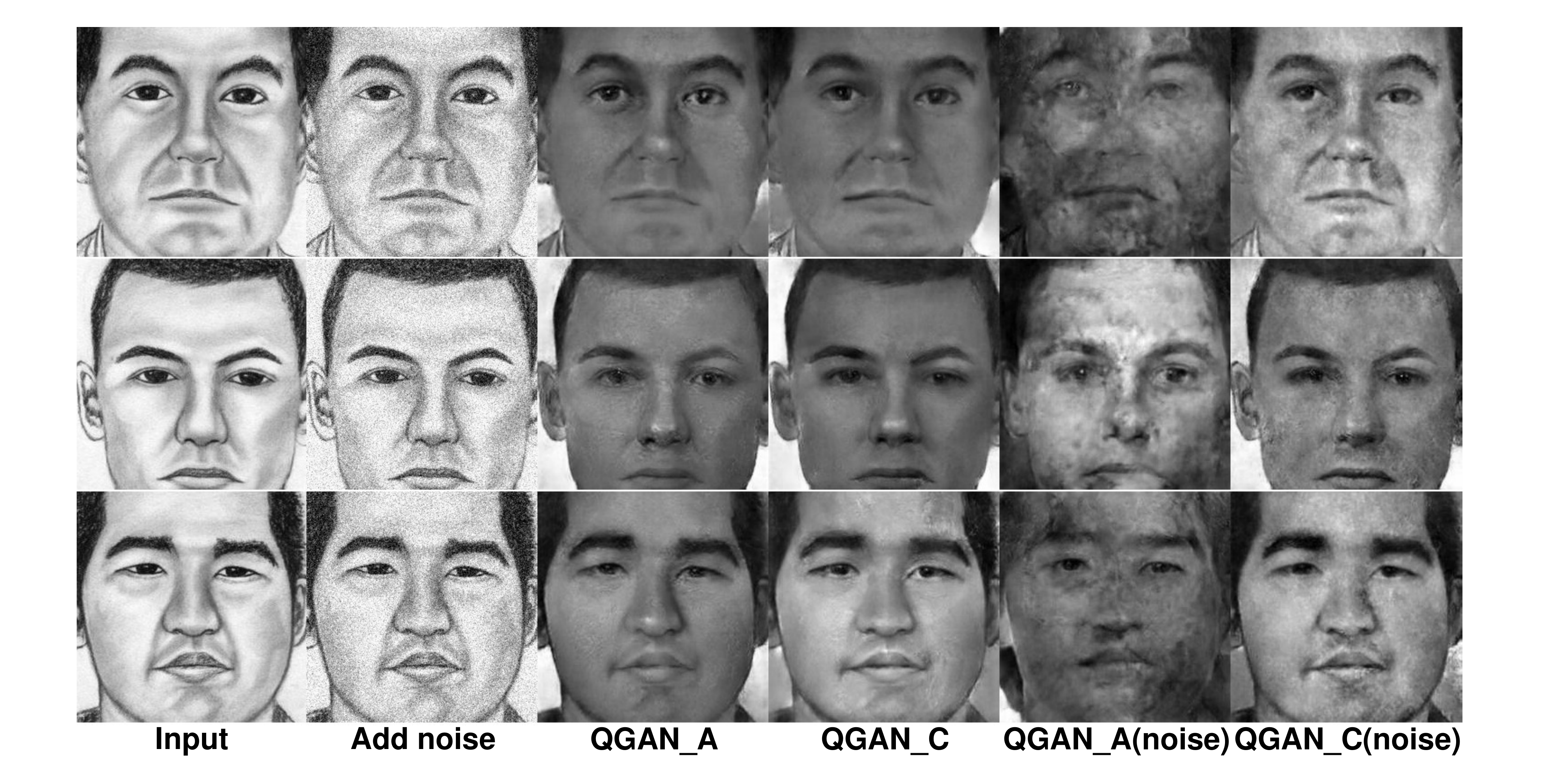}
\vspace{-0.3cm}
\caption{Typical failure cases of QGAN\_A. ``Add noise" represents we add Gaussian noise~(mean=0, var=0.001) to the original images.}
\label{Fig:failcase}
\end{center}
\vspace{-0.8cm}
\end{figure}

For most experimental cases, the results of QGAN\_A outperform QGAN\_C. However, as discussed before, the QGAN\_A  would fail if the training data contains noise. To empirically validate this, we perform an experiment as shown in Fig.~\ref{Fig:failcase}. In this figure,  we add Gaussian noise~(mean=0, var=0.001) to inputs for the sketch$\rightarrow$photo translation task. The last two columns are the generated images with the noisy inputs. We observe QGAN\_A performs much worse than QGAN\_C. The reason is that: the FSIM based loss contains the  phase congruency measure, which is sensitive to noises~\cite{kovesi2000phase}. In contrast, the denoising effect is automatically achieved by the ``U-NET'' architecture of generators in QGAN\_C.



\section{Conclusion}
In this paper, we revisited the problem of unpaired image-to-image translation, and designed a unified QGAN framework
for quality-aware unpaired image translation. In the QGAN framework, a quality-aware loss term is explicitly incorporated in the optimization function.  Specifically, we designed two detailed implementations of the quality loss, i.e., QGAN\_A and QGAN\_C, that considered the classical quality assessment model and the adaptive high level content structure information from deep network. Extensive quantitative comparisons against prior models, as well as a mean opinion score test clearly showed the superior quality of our proposed framework and the two detailed implementations.
In the future, we would like to apply our proposed framework to image applications that rely on image quality, e.g., image super resolution.

%
%

\bibliographystyle{IEEEtran}
\bibliography{paper_bib}

\begin{IEEEbiography}[{\includegraphics[width=1in,height=1.25in,clip,keepaspectratio]{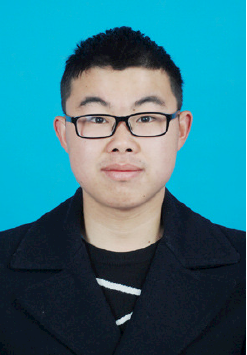}}]{Lei Chen}
is currently working towards the M.S. degree at Hefei University of Technology, China. He received the B.S. degree from Anhui University in 2016. His research interests include multimedia analysis and data mining.
\end{IEEEbiography}

\begin{IEEEbiography}[{\includegraphics[width=1in,height=1.25in,clip,keepaspectratio]{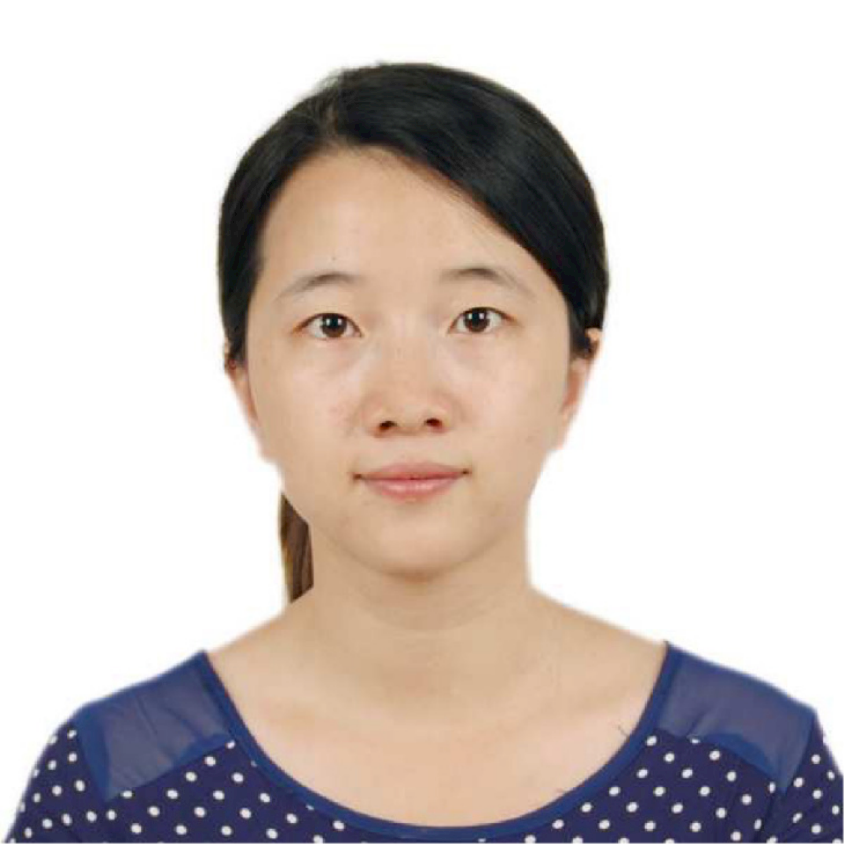}}]{Le Wu}
is currently an assistant professor at the Hefei University of Technology (HFUT), China. She received the Ph.D. degree from the University of Science and Technology of China (USTC). Her general area of research  interests is data mining, recommender systems and social network analysis. She has published more than 30 papers in referred journals and conferences. Dr. Le Wu is the recipient of the Best of SDM 2015 Award, and the Distinguished Dissertation Award from China Association for Artificial Intelligence (CAAI) 2017.
\end{IEEEbiography}

\begin{IEEEbiography}[{\includegraphics[width=1in,height=1.25in,clip,keepaspectratio]{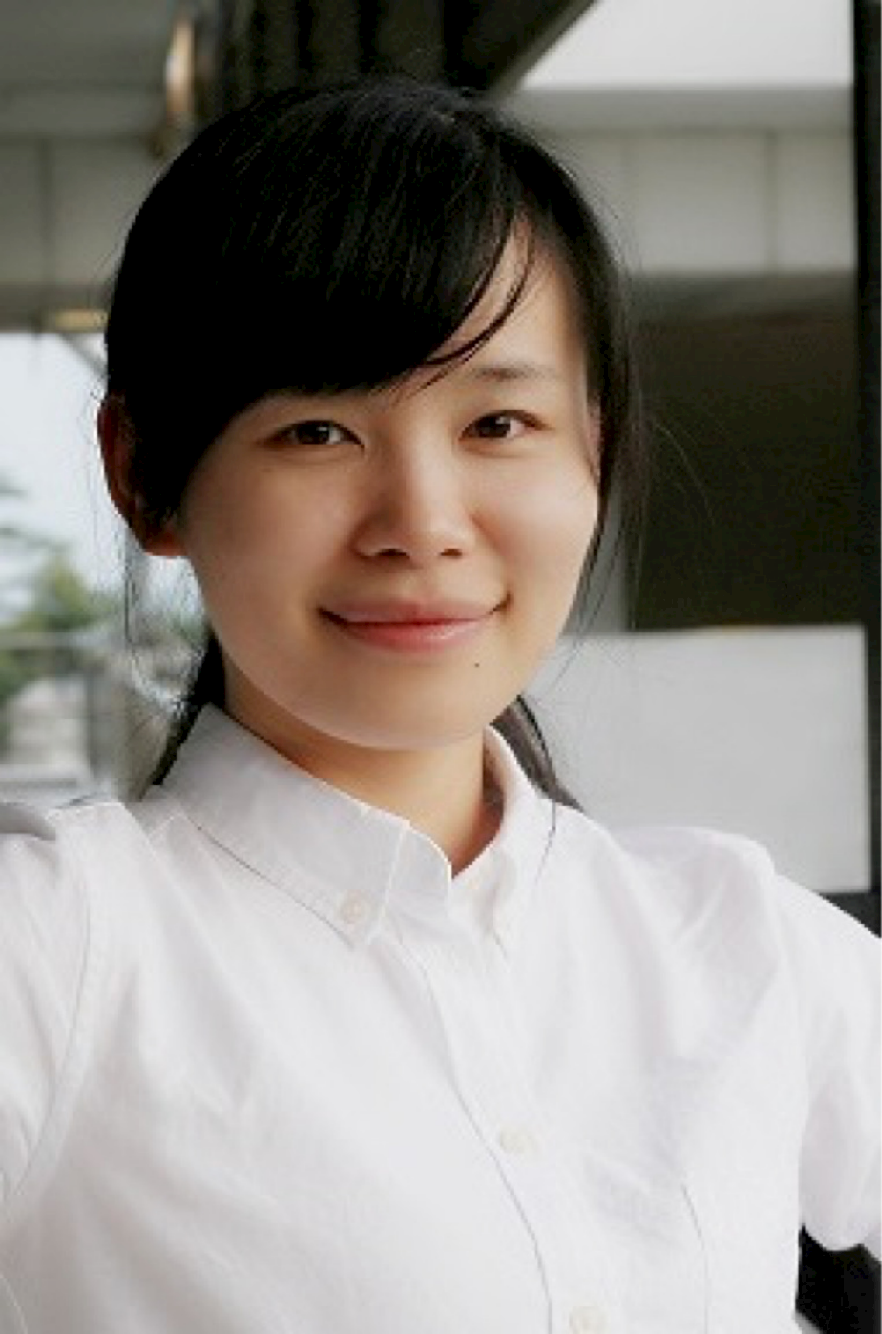}}] {Zhenzhen Hu} is an associate professor with School of Computer and Information at Hefei University of Technology (HFUT), China. She received her PhD degree from HFUT in 2014, under the supervision of Prof. Jianguo Jiang and Prof. Richang Hong. She used to be a Research Fellow at Nanyang Technological University, directed by Prof. Yonggang Wen. Her current research interests include cross-media computing and computer vision.
\end{IEEEbiography}

\begin{IEEEbiography}[{\includegraphics[width=1in,height=1.25in,clip,keepaspectratio]{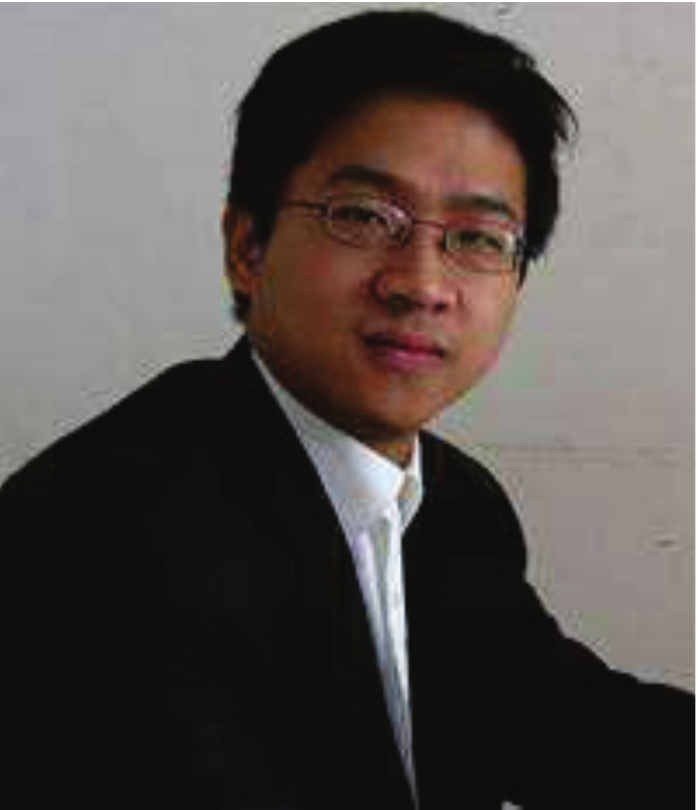}}]{Meng Wang } is a professor at the Hefei University of Technology,
China. He received his B.E. degree and Ph.D. degree in the Special
Class for the Gifted Young and the Department of Electronic
Engineering and Information Science from the University of Science and
Technology of China (USTC), Hefei, China, in 2003 and 2008,
respectively. His current research interests include multimedia
content analysis, computer vision, and pattern recognition. He has
authored more than 200 book chapters, journal and conference papers in
these areas. He is the recipient of the ACM SIGMM Rising Star Award 2014.
He is an associate editor of IEEE Transactions on Knowledge and Data
Engineering (IEEE TKDE), IEEE Transactions on Circuits and Systems
for Video Technology (IEEE TCSVT), IEEE Transactions on Multimedia (IEEE TMM), and IEEE Transactions on Neural Networks and Learning Systems (IEEE TNNLS).
\end{IEEEbiography}

\end{document}